\pdfoutput=1

\documentclass[11pt]{article}

\usepackage[final]{acl}

\usepackage{times}
\usepackage{latexsym}
\usepackage{booktabs}
\usepackage{tcolorbox}

\usepackage[T1]{fontenc}

\usepackage[utf8]{inputenc}

\usepackage{microtype}

\usepackage{inconsolata}

\title{Intrinsic Task-based Evaluation for Referring Expression Generation}

\author{Guanyi Chen$^{\clubsuit}$, Fahime Same\textsuperscript{$\heartsuit$}, Kees van Deemter\textsuperscript{$\spadesuit$}\\
  $^{\clubsuit}$Hubei Provincial Key Laboratory of Artificial Intelligence and Smart Learning, \\
  National Language Resources Monitoring and Research Center for Network Media, \\
  School of Computer Science, Central China Normal University \\
  \textsuperscript{$\heartsuit$}Department of Linguistics, University of Cologne\\
\textsuperscript{$\spadesuit$}Department of Information and Computing Sciences, Utrecht University\\
  \texttt{g.chen@ccnu.edu.cn, f.same@uni-koeln.de, c.j.vandeemter@uu.nl}}

\begin{document}
\maketitle
\begin{abstract}
Recently, a human evaluation study of Referring Expression Generation (REG) models had an unexpected conclusion: on \textsc{webnlg}, Referring Expressions (REs) generated by the state-of-the-art neural models were not only indistinguishable from the REs in \textsc{webnlg} but also from the REs generated by a simple rule-based system. 
Here, we argue that this limitation could stem from the use of a purely ratings-based human evaluation (which is a common practice in Natural Language Generation). To investigate these issues, we propose an intrinsic task-based evaluation for REG models, in which, in addition to rating the quality of REs, participants were asked to accomplish two meta-level tasks. One of these tasks concerns the referential success of each RE; the other task asks participants to suggest a better alternative for each RE. The outcomes suggest that, in comparison to previous evaluations, the new evaluation protocol assesses the performance of each REG model more comprehensively and makes the participants' ratings more reliable and discriminable.
\end{abstract}


\section{Introduction} \label{sec:intro}

Referring Expression Generation (REG) is a key aspect of Natural Language Generation (NLG) and a vital step in the classic Natural Language Generation pipeline~\citep[NLG,][]{reiter_dale_2000,10.5555/3241691.3241693}. REG produces referring expressions (REs) that refer to referents at different points in a discourse~\citep{belz-varges-2007-generation}. It has great practical value for commercial NLG~\citep{reiter-2017-commercial} and is actively studied by theoretical linguists and psycholinguists~\citep{van2016computational}. 

Conventional REG systems have two steps. The first step determines the rough form of the RE. For instance, given a certain context, the system needs to decide whether a given reference to Homer Simpson should be a proper name, a pronoun, or a description. The second step involves 
choosing a concrete Noun Phrase. For example, if step 1 says a proper name must be chosen, then the second step involves choosing between ``\emph{Homer}'' or ``\emph{Homer Simpson}''; if a description must be chosen, then many options exist, including  ``\emph{the protagonist of the American animated sitcom The Simpsons}''. 

\citet{castro-ferreira-etal-2018-neuralreg} redefined the task of REG on the WebNLG corpus~\citep[][which has been widely used in NLG research]{gardent-etal-2017-creating,castro-ferreira-etal-2018-enriching} and tackled it in an End2End manner, i.e., tackling the two above-mentioned steps simultaneously, using Deep Neural Networks. Later, many follow-ups were carried out to strengthen the ``NeuralREG'' model (e.g.,~\citet{cao-cheung-2019-referring, cunha-etal-2020-referring}). 

Nonetheless, in a large-scale human evaluation of NeuralREG models, \citet{same-etal-2022-non} surprisingly found that, for human readers, the REs generated by NeuralREG models are not only indistinguishable from the REs in a corpus but also from the REs generated by a simple rule-based model. For this reason, they questioned the usefulness of neural models in REG, as well as the suitability of the WebNLG corpus as a testing ground for the strengths of the models.

An alternative take on these counter-intuitive results is that the method of asking participants to rate models' outputs 
may not have been sufficiently sensitive.
As previous studies have pointed out~\citep{THOMSON2023101482,goyal-etal-2023-factual}, pragmatic appropriateness is difficult to assess especially for REs, since the use of REs is highly context-dependent and diverse. Note that previous human evaluations for REG models~\citep{castro-ferreira-etal-2018-neuralreg,cunha-etal-2020-referring,same-etal-2022-non} always asked participants to rate 
an entire discourse, which may have failed to make participants aware of the use of REs (even though their instructions asked them to focus on REs). 

\begin{table*}[t!]
\small
\centering
\begin{tabular}{p{15cm}}
\toprule
\textbf{Triples}: \\
AWH\_Engineering\_College | country | India \\
Kerala | leaderName | Kochi \\
AWH\_Engineering\_College | academicStaffSize | 250 \\
AWH\_Engineering\_College | state | Kerala \\
AWH\_Engineering\_College | city | ``Kuttikkattoor'' \\
India | river | Ganges \\
\midrule
\textbf{Text}: AWH Engineering College is in Kuttikkattoor, India in the state of Kerala. The school has 250 employees and Kerala is ruled by Kochi. The Ganges River is also found in India. \\
\textbf{Delexicialised Text}: \textcolor{blue}{\underline{AWH\_Engineering\_College} is in \underline{``Kuttikkattoor''} , \underline{India} in the state of \underline{Kerala} . }\textbf{\underline{AWH\_Engineering\_College}} \textcolor{red}{has 250 employees and \underline{Kerala} is ruled by \underline{Kochi} . \underline{Ganges} is also found in \underline{India} .}\\
\bottomrule
\end{tabular}
\caption{An example data from the \textsc{webnlg} corpus. In the delexicalised text, every entity is \underline{underlined}, the target entity is \textbf{boldfaced}, the pre-context is coloured in \textcolor{blue}{blue}, and the post-context is coloured in \textcolor{red}{red}. The upper part of the table shows the 6 predicate-argument relations; therefore, the size of the triple is 6.}
\label{tab:sample}
\end{table*}

In this paper, we use Intrinsic Task-based Human Evaluation. In addition to rating each discourse, participants were asked to accomplish two meta-level tasks.
One task is judging the \textit{referential success of each RE}, i.e., determining whether the RE indeed refers to its intended target referent. The other involves \textit{optimising the REs} that, in their opinion, are not optimal, to enhance the clarity and coherence of the discourse.

We believe that this new approach to evaluation has much to add to existing approaches because (1) answering the questions about REs and optimising REs by hand helps participants zoom in on the use of REs in each discourse during the human evaluation, making their score more reliable and discriminable; (2) the participants' responses allow us to evaluate REG models in a more comprehensive way and from multiple angles; and (3) On the basis of rewritings from participants, we can gain further insights about human language use. 

In the current study, based on the above protocol, we carried out a human evaluation experiment on the models' outputs from \citet{same-etal-2022-non}. Building on participants' responses, we first analysed the performance of each REG model comprehensively; we then compared our results to those of~\citet{same-etal-2022-non} to see how accomplishing the above-mentioned tasks has impacted participants' rating behaviour. Finally, we looked into the contents of these responses and summarised our observations. 
\section{Background} \label{sec:background}

First, we explain the REG task and the corpus it is based on. Then we outline
the experiments of~\citet{same-etal-2022-non}. 

\subsection{The REG task}

Given a text whose REs are not yet realised, the  REG task is to 
generate the REs that refer to their intended referents.
This definition was further elaborated by~\citet{castro-ferreira-etal-2018-neuralreg} based on the \textsc{webnlg} corpus.

\textsc{webnlg} was constructed by asking crowdworkers to write descriptions for a set of Resource Description Framework (RDF) triples (Table~\ref{tab:sample}); each entity is represented by its delexicalised proper name (e.g., ``AWH\_Engineering\_College'' for the entity AWH Engineering College); we call this representation the \emph{identifier} of the entity as it is unique in the corpus. The number of triples varied from 1 to 7. To fit the REG task, \citet{castro-ferreira-etal-2018-neuralreg} used triples to delexicalise each description (see the delexicalised text in Table~\ref{tab:sample}) and trained a model to generate the surface form of the entity (i.e., \emph{The school}) given its identifier
(i.e., ``AWH\_Engineering\_College''), its pre-context (``\emph{AWH\_Engineering\_College is in ``Kuttikkattoor'' , India in the state of Kerala . }''), and its post-context (``\emph{has 250 employees and Kerala is ruled by Kochi . The Ganges River is also found in India.}''). 
To test the generalizability of REG models, \citet{castro-ferreira-etal-2018-neuralreg} divided the test set of \textsc{webnlg} into \emph{seen} set (where all data are from the same domains as the training data) and \emph{unseen} set (where all data are from different domains than the training data).

\subsection{The Results in \citet{same-etal-2022-non}}

Based on  the \textsc{webnlg} corpus and the REG task defined by~\citet{castro-ferreira-etal-2018-neuralreg}, \citet{same-etal-2022-non} 
evaluated several state-of-the-art (SOTA) neural REG models, including \texttt{ATT+Copy}~\citep{castro-ferreira-etal-2018-neuralreg}, \texttt{ATT+Meta}~\citep{cunha-etal-2020-referring} and \texttt{ProfileREG}~\citep{cao-cheung-2019-referring}. Additionally, they 
tested two rule-based models, one with a small set of rules (\texttt{RREG-S}) and one with a larger set of rules (\texttt{RREG-L}); as well as two machine learning-based (ML) models, one with a small set of features (\texttt{ML-S}) and one with a large set of linguistically motivated features (\texttt{ML-L}).\footnote{Since the underlying mechanisms of these models are not the focus of the present paper, we skip the introduction of their details. Please check \citet{same-etal-2022-non} for details.}

Counter-intuitively, \citeauthor{same-etal-2022-non} found out that in the case of WebNLG, participants found all models to be almost indistinguishable from each other and from the REs in the original \textsc{webnlg} documents in terms of clarity, grammaticality and coherence. 
They hypothesised that this might be because the data in \textsc{webnlg} does not accurately reflect the everyday use of REs. Consequently, they built a new REG dataset based on the Wall Street Journal (\textsc{wsj}) portion of the OntoNotes corpus \citep{weischedel2013ontonotes}. More surprisingly, assessments on \textsc{wsj} indicated that both the simple rule-based (\texttt{RREG-S}) model and the linguistically-informed ML-based (\texttt{ML-L}) model significantly outperformed the two more advanced neural REG models. 

These unexpected results, coupled with the lack of distinguishable statistically significant differences in the case of the \textsc{webnlg} models, motivated us to perform a more in-depth evaluation of the model outcomes. We begin this evaluation with the simpler corpus, namely, \textsc{webnlg}, and plan to expand it in the future to the more complex dataset.
%

\section{Research Questions} \label{sec:rq}

As motivated in the introduction, our enhanced experiment rests on three key tasks. The first of these was the same ratings-based task that had been used in earlier studies (including~\citealt{same-etal-2022-non}), namely to rate the clarity, grammaticality and coherence of the item in which a given coreference chain occurs (see Figure~\ref{fig:example}). The second task was to judge the referential success of the REs. The third was for participants to suggest a rewriting for each RE wherever they felt this to be necessary.\footnote{Note that Similar ideas were used in the extrinsic evaluation of practical NLG systems (in contrast to our intrinsic evaluation), such as~\citet{sripada-etal-2005-evaluation}, in which post-edits were used to understand how experts think about the outputs of a weather forecast generation system.} Our principal aim was to find out how the outcomes of the new experiments differ from earlier experiments; in particular, we were curious \emph{whether the indistinguishability results of \citet{same-etal-2022-non} may have been caused by the limitations of a human evaluation method that is based solely on ratings.} 

Since our experiment produced three different ways of assessing the quality of an RE (namely its rating, its referential success, and the frequency of rewriting), the question comes up about \emph{how these three assessments are related to each other}: do they tend to point in the same direction, or not? Note that even if these assessments proved to be very closely aligned, it would not follow that two of the three are superfluous, because the fact that this plurality of tasks forced participants to immerse themselves deeply in the texts may have improved the quality of all their responses, including those for the familiar ratings-based task.

Finally, by studying the corpus, it occurred to us that referents differed sharply from each other along a potentially important dimension, namely whether a participant was previously familiar with the referent or not. It has been pointed out that, in some settings, both speakers and hearers use/interpret REs differently if they are familiar with the  referent~\citep{kutlak2011audience,staliunaite-etal-2018-getting}. 
We, therefore, asked participants, for each referent, whether or not they were familiar with this referent before they read the text, and we investigated \emph{how participants' familiarity with a referent affected their responses on the key tasks}.
%

Based on the assumptions mentioned above and the questions raised by these assumptions, we put forward the following hypotheses: 

\begin{description}
\item[Hypothesis 1 ($\mathcal{H}_1$)] The meta-level tasks (i.e., the referential success task and rewritings) help participants make more informed ratings. Therefore, unlike the findings by~\citet{same-etal-2022-non}, we expect to observe significant distinguishable differences in the ratings of the models.
\item[Hypothesis 2 ($\mathcal{H}_2$)] Since the tasks help participants identify inappropriate REs in each text, we expect that the scores in this experiment would be lower than those in~\citet{same-etal-2022-non}.
\item[Hypothesis 3 ($\mathcal{H}_3$)] Regarding the referential success of REs, we expect that the more frequently REs are marked as successful, the higher the discourse would be rated.
\item[Hypothesis 4 ($\mathcal{H}_4$)] We expect that the more often the REs are re-written, the lower the scores the discourse would receive. 
\item[Hypothesis 5 ($\mathcal{H}_5$)] We expect that participants would spot more inappropriate REs in discourse if they are familiar with its major entity and, as a consequence, they would re-write more and rate the discourse lower.
\end{description}

\section{Intrinsic Task-based Human Evaluation} \label{sec:method}

\begin{figure*}[t!]
    \begin{tcolorbox}[colback = white, boxrule = 0.3mm]
    {\small
    \emph{AWH Engineering College is in Kuttikkattoor, India in the state of Kerala. The school has 250 employees and Kerala is ruled by Kochi. The Ganges River is also found in India.} \\[5px]
    \underline{Question 1}: The above text is a short introduction to Adirondack Regional Airport. Were you aware of the existence of AWH Engineering College before this experiment? [Yes, No]\\[5px]
    \underline{Question 2}: Regarding the above paragraph, please answer the question(s) below.\\
    1. Does the expression ``The school'' (highlighted) in the text below refer to ``AWH Engineering Colleg''? [Yes, Maybe, No]\\
    \emph{AWH Engineering College is in Kuttikkattoor, India in the state of Kerala. \textbf{The school} has 250 employees and Kerala is ruled by Kochi. The Ganges River is also found in India.} \\
    2. (...) \\[5px]
    \underline{Question 3}: We have highlighted some of the expressions in the paragraph. Please focus on these expressions when answering the questions below.\\
    \emph{\textbf{AWH Engineering College} is in \textbf{Kuttikkattoor}, \textbf{India} in the state of \textbf{Kerala}. \textbf{The school} has 250 employees and \textbf{Kerala} is ruled by \textbf{Kochi}. \textbf{The Ganges River} is also found in \textbf{India}.}\\
    Focusing on these expressions, to what extent do you think the following statements are true?\\
    - This paragraph is clear. [1-7]\\
    - This paragraph is grammatical. [1-7]\\
    - This paragraph is coherent. [1-7]\\
    Below, we have listed these expressions in the paragraph. Could you suggest a better alternative for each expression to enhance the paragraph's coherence, grammatical correctness, and clarity? For the one that you think is optimal, you can simply copy and paste the expression in the paragraph into the text box. (For your convenience, we show you the highlighted paragraph again with each expression numbered).\\
    \emph{\textbf{(1) AWH Engineering College} is in \textbf{(2) Kuttikkattoor}, \textbf{(3) India} in the state of \textbf{(4) Kerala}. \textbf{(5) The school} has 250 employees and \textbf{(6) Kerala} is ruled by \textbf{(7) Kochi}. \textbf{(8) The Ganges River} is also found in \textbf{(9) India}.} \\
    - (1) AWH Engineering College: \\
    - (2) (...) \\[5px]
    Apart from these expressions, do you have any other comments or suggestions? (Optional)
    }
    \end{tcolorbox}
    \caption{An example item in our experiment.}
\label{fig:example}
\end{figure*}

Here we describe the setup of the new experiment.
We consider this to be an ``intrinsic'' NLG evaluation, as opposed to the non-real-world (``extrinsic'') tasks more commonly associated with task-based NLG evaluation \cite{reiter2011task}. 

\subsection{Materials}

We used the same set of items from the human evaluation experiment of~\citet{same-etal-2022-non}, in which there are seen data and unseen data sampled from the test set of \textsc{webnlg}.\footnote{The data from~\citet{same-etal-2022-non} is available at \url{https://github.com/a-quei/neuralreg-re-evaluation}.} More specifically, there are 48 seen items (4 items from each triple size group of 2-7) and 48 unseen items (6 items from each triple size group of 2-5). 
In this study, we considered 5 REG models. Based on models' performance on both \textsc{webnlg} and \textsc{wsj} reported in \citet{same-etal-2022-non}, we selected the two best-performing neural models (\texttt{ATT+Copy} and \texttt{ATT+Meta}, henceforth, \texttt{NREG-1} and \texttt{NREG-2}, respectively, since the specific details of each model are not our focus), the best rule-based model (\texttt{RREG-S}, henceforth, \texttt{RULE}), and the best ML-based model (\texttt{ML-L}, henceforth, \texttt{ML})\footnote{The full results of these models from~\citet{same-etal-2022-non} can be found in Appendix~\ref{sec:appendix_a}}. In other words, for each test item, in addition to the reference text, we used 4 versions generated by the models, resulting in a total number of 240 test items ($48 \times 5$).

\subsection{Experiment Design}

The 240 items were divided into 16 groups of 15 items each through a pseudo-randomisation process. This process ensured, to the greatest extent possible, that two versions of the same test item did not appear in the same group. 

At the beginning of the experiment, we explained the 
goal of the experiment to the participants in broad terms and we clarified what we expected them to do. The full instruction can be found in Appendix~\ref{sec:instruction}.

Since our participants were not linguistic experts, we opted for simpler terminology, using ``expression'' instead of ``referring expression'' and ``paragraph'' instead of ``text'' or ``discourse''. A slight drawback of this formulation is that it does not forbid the use of phrases that are not REs or even noun phrases when suggesting rewritings. In fact, however, since the non-REs were very infrequent in the outcomes and we felt that this should not affect the testing of our hypotheses, we processed all rewritings in the same way (see Section~\ref{sec:data}).

Each item contains several questions and tasks that participants need to accomplish, an example of which is shown in Figure~\ref{fig:example}. It starts by showing participants the whole item's text, which is followed by the following questions/tasks.\footnote{Henceforth, the term ``text" will refer to any one of the short paragraphs that we presented to participants and that can be understood by itself.}

\paragraph{Familiarity.} 
The \textsc{webnlg} dataset contains data units that are composed of RDF triples, each extracted from DBPedia. The accompanying texts for these data units are sequences of one or more sentences that verbalise the information in the RDF triples \citep{gardent-etal-2017-webnlg}. These texts revolve around a central entity.
We call this entity the \emph{major entity} in the discourse. The first question in the experiment is a Yes-No question, asking whether the participant is familiar with the major entity. 
In the question, we referred to the major entity using its proper name, which is obtained by replacing underscores in its identifier (hereafter, \textsc{proper name}).

\paragraph{Referential Success.}
The second question asks whether the RE in question is successful in identifying the referent.  
This question was asked only for REs that differ from their \textsc{proper name}. 
For instance, in the example in Table~\ref{tab:sample}, the RE ``\emph{the school}'' is one such case. This expression differs from the proper name format of the identifier. The identifier, in this case, is ``AWH\_Engineering\_College'', and its \textsc{proper name} is ``AWH Engineering College''. We highlighted the RE and asked whether the expression ``\emph{the school}'' refers to ``AWH Engineering College".

Moreover, in our pilot study, we found that participants were sometimes unsure whether an RE was successful or not. Thus, we added a ``Maybe'' option. This option could also provide us with insights about the REs that are more difficult to resolve.

It is worth noting that \textsc{webnlg} contains a few errors. For example, it occasionally marks ``American'' as referring to the United States. We manually corrected these errors while preparing the experiment materials.

\paragraph{Rating.} 

Given our aim to perform a quantitative evaluation of REG models and our curiosity about the impact of the meta-linguistic tasks on the overall scores, we asked participants (in the first part of the third question, see Figure~\ref{fig:example}) to rate the text in the same manner as described in~\citet{same-etal-2022-non}. Concretely, participants were asked to answer whether they agreed with the following three statements on a 7-point Likert-scale, where 1 means ``strongly disagree'', 4 means ``I don't know'', and 7 means ``strongly agree'': (1) Clarity: This paragraph is clear; (2) Grammaticality: This paragraph is grammatical; and (3) Coherence: This paragraph is coherent. Since factors other than the quality of REs can also influence the overall quality of a text, we asked participants to rate the texts while focusing on the REs highlighted in the discourse.

\paragraph{Rewriting.} As discussed in Section~\ref{sec:rq}, finally, we asked participants to suggest better rewritings in the second part of question 3 (see the example in Figure~\ref{fig:example}). Participants were instructed to ``\emph{suggest a better alternative for each expression to enhance the paragraph's coherence}''. To make sure that no RE was overlooked or skipped, we made it mandatory for them to write a suggestion for each RE. We mentioned that if they found an RE to be optimal, they could simply copy and paste the original RE into the designated slot. 

It is worth noting that we placed the rewriting task after the rating task because we expected that completing the rating first would lead the participants to read the entire text before beginning to optimise its REs. We combined these two tasks into a single question with the expectation that participants might re-evaluate their scores following the rewriting task.   

\paragraph{Additional Comments.} 

During the pilot study, we observed that some participants criticised non-referential aspects of the
text  (i.e., aspects that did not concern the REs in the text).
Therefore, in the main experiment, we allowed participants to provide comments on aspects other than just REs. 

\subsection{Participants and Procedure}

We constructed the experiment materials using Qualtrics and carried out the experiment on Prolific. Each set of items was completed by 8 participants. We restricted participants to be native speakers of English and to those located in the United Kingdom or the United States. Based on the pilot study, we expected that participants would complete the experiment, which consisted of 15 items, within 30 minutes. Therefore, we paid Proflic 5 pounds for each participant. We rejected participants if they (1) apparently misunderstood any of our tasks, or (2) wrote nonsensical responses.

Besides the demographic information available on prolific, we did not collect any additional personal information during the experiment. 

We obtained responses from 128 participants ($16 \times 8$), with an average age of 38. Of these, 75 identified as female, and 51 identified as male. The average duration of the experiment was 41 minutes, which was higher than what we expected. This was because several participants had strong views on the use of RE. They tried to optimise every RE and left very long comments, which made them spend more than an hour on our experiment.
\section{Results} \label{sec:result}

In this section, we introduce the data we obtained from the experiment, explain how we post-processed the data, and report the results. 

\subsection{The Dataset} \label{sec:data}

There are a total of 1325 REs in the 240 test items. Out of these, 469 REs are different from their \textsc{proper name}s. We asked about the referential success of these REs and received 3752 responses~($469 \times 8$). 

For these 1325 REs, we obtained 10600 participant-written REs from our participants. These REs may contain typos or formatting issues; for example, some participants wrote short comments in the text boxes intended for writing suggested REs. Therefore, we manually corrected every participant-written RE and annotated whether the RE was a rewriting or a copy of the original RE. Ultimately, we obtained 2832 rewritings.

It is worth noting that, during annotation, we found that some participants commented on certain REs that ``\emph{it is impossible to infer which referent this RE refer to}'' or ``\emph{this RE is redundant}''. We annotated the former case as ``unresolvable'' and the latter case as ``redundant''. Out of 10600 participant-written REs, we identified 53 ``unresolvable'' cases for 48 REs (an RE can be marked ``unresolvable'' by multiple participants) and 39 ``redundant'' cases for 17 REs. We treated these cases as rewritings. The data will be 
available at: \url{https://github.com/a-quei/reg-rewriting}. 


 \begin{table*}[t!]
\small
\centering
\begin{tabular}{lcccccccccccc}
\toprule
 & \multicolumn{4}{c}{All} & \multicolumn{4}{c}{Seen} & \multicolumn{4}{c}{Unseen} \\\cmidrule(lr){2-5} \cmidrule(lr){6-9}\cmidrule(lr){10-13}
Model & SR & Yes & Maybe & No & SR & Yes & Maybe & No & SR & Yes & Maybe & No \\ \midrule
\texttt{RULE} & 99.39 & 90.97 & 6.25 & 2.68 & 99.39 & 90.28 & 5.56 & 4.17 & 99.39 & 91.67 & 6.94 & 1.39 \\
\texttt{ML} & 94.14 & 77.01 & 17.63 & 5.36 & 91.08 & 75.94 & 18.40 & 5.66 & 99.90 & 95.83 & 4.17 & 0.00 \\
\texttt{NREG-1} & 80.80 & 64.67 & 25.87 & 9.46 & 93.18 & 83.75 & 11.67 & 4.58 & 66.29 & 51.04 & 36.01 & 12.95 \\
\texttt{NREG-2} & 81.51& 64.23 & 25.82 & 9.95 & 90.21 & 78.12 & 15.23 & 6.64 & 71.31 & 52.05 & 35.10 & 12.84 \\ \midrule
\texttt{Human} & 94.62 & 87.50 & 9.32 & 3.18 & 93.79 & 85.45 & 10.04 & 4.51 & 95.59 & 89.86 & 8.49 & 1.65 \\
\bottomrule
\end{tabular}
\caption{The proportion of each response to the questions concerning the referential success of REs. SR stands for successful rate of REs.}
\label{tab:success_webnlg}
\end{table*}

\subsection{Main Results}

\paragraph{Referential Success.}

Table~\ref{tab:success_webnlg} presents the results for the answers to questions that ask about the referential success of REs. It includes the proportion of each response type (`Yes', `Maybe', and `No') for each model. Additionally, we computed the Success Rate (SR), which is defined as the number of `Yes' responses a model received to the product of the number of REs and the number of participants. SR considers \textsc{proper name}s (see Section~\ref{sec:method}) as referentially successful REs. These numbers for Seen and Unseen portions are also reported. The raw count of each answer can be found in Appendix~\ref{sec:appendix_b}.

\texttt{RULE} has the highest SR among all other models, including \texttt{Human}. This is because it used \textsc{proper name}s in the majority of cases and, for the rest, chose only pronouns. By never using descriptions, the REs it generated were easier to resolve. The SR of \texttt{ML} is on par with \texttt{Human}. This model works remarkably well on unseen data.\footnote{Recall that some models work better on unseen data, primarily because this data is simpler than seen data. See Section~\ref{sec:background} for more details.}
None of the REs it generated for this data were ever marked as being definitely unsuccessful. On the seen data, the REs generated by \texttt{ML} and marked as ``maybe'' were often pronouns with slight referential ambiguity.
Compared to \texttt{RULE} and \texttt{ML}, the two neural models were more likely to produce ambiguous REs. Nonetheless, the REs they generated for seen data are equally as successful as \texttt{Human}, but those for unseen data are dramatically worse.  

\paragraph{Ratings.} 
\begin{table*}[t!]
\small
\centering
\begin{tabular}{lcccccccccccc}
\toprule
 & \multicolumn{4}{c}{All} & \multicolumn{4}{c}{Seen} & \multicolumn{4}{c}{Unseen} \\\cmidrule(lr){2-5} \cmidrule(lr){6-9}\cmidrule(lr){10-13}
Model & C & G & Co & RR & C & G & Co & RR & C & G & Co & RR \\ \midrule
\texttt{RULE} & 4.78$^{B}$~~~ & 4.12$^{B}$ & 4.62$^{B}$ & 28.44 & 4.61$^{A}$ & 3.84$^{B}$~~~ & 4.33$^{B}$~~~ & 29.81 & 4.95$^{A}$ & 4.40$^{B}$ & 4.92$^{A}$ & 26.84 \\
\texttt{ML} & 4.79$^{B}$~~~ & 4.34$^{B}$ & 4.63$^{B}$ & 26.51 & 4.61$^{A}$ & 4.28$^{A,B}$ & 4.45$^{B}$~~~ & 26.05 & 4.98$^{A}$ & 4.40$^{B}$ & 4.81$^{A}$ & 27.05 \\
\texttt{NREG-1} & 4.49$^{B,C}$ & 4.18$^{B}$ & 4.36$^{B}$ & 27.31 & 5.01$^{A}$ & 4.51$^{A}$~~~ & 4.81$^{A,B}$ & 20.37 & 3.98$^{B}$ & 3.86$^{B}$ & 3.91$^{B}$ & 35.45 \\
\texttt{NREG-2} & 4.40$^{C}$~~~ & 4.15$^{B}$ & 4.33$^{B}$ & 30.90 & 4.79$^{A}$ & 4.43$^{A}$~~~ & 4.70$^{A,B}$ & 24.30 & 4.01$^{B}$ & 3.88$^{B}$ & 3.96$^{B}$ & 38.63 \\ \midrule
\texttt{Human} & 5.17$^{A}$~~~ & 4.85$^{A}$ & 5.10$^{A}$ & 20.42 & 5.05$^{A}$ & 4.63$^{A}$~~~ & 4.97$^{A}$~~~ & 21.33 & 5.30$^{A}$ & 5.06$^{A}$ & 5.22$^{A}$ & 19.36 \\
\bottomrule
\end{tabular}
\caption{The rating results as well as re-writing rates (RR). `C`, `G' and `Co' stand for Clarity, Grammaticality, and Coherence, respectively. Rankings are determined by significance testing ($p < 0.01$; using Wilcoxon's signed-rank test with Bonferroni correction). Per column, results that have \emph{no} superscript letters in common are significantly different from each other. Note that, the lower the RR, the better.}
\label{tab:rating_webnlg}
\end{table*}

Table~\ref{tab:rating_webnlg} reports the participants' ratings. Compared to \citet{same-etal-2022-non} (cf. Table~\ref{tab:human_webnlg_2022} in Appendix~\ref{sec:appendix_a}), on the same set of test samples, scores from our experiment are significantly lower (in terms of clarity, grammaticality and coherence using a Mann-Whitney Test; $p<.001$), which confirms our hypothesis $\mathcal{H}_2$. 

Unlike ~\citet{same-etal-2022-non}, \texttt{Human} (the original texts) achieves significantly better performance than all the other models in terms of all three criteria (using Wilcoxon’s signed-rank
test with Bonferroni correction). The experimental models seem to be still indistinguishable from each other except \texttt{NREG-2}, which has the lowest clarity score. 

Zooming in on the seen data, the two neural models perform equally well to \texttt{Human} while non-neural models (especially \texttt{RULE}) receive lower grammaticality and coherence scores. On unseen data, the situation is the other way around. Since unseen data has lower complexity, \texttt{RULE} and \texttt{ML} can produce clear and coherent REs, but the grammaticality of these REs is still a problem. Meanwhile, neural models are significantly worse than other models and \texttt{Human}. In short, we observe significant differences in ratings, confirming $\mathcal{H}_1$.

Moreover, these findings are consistent with the analysis of the referential success of REs, revealing that non-neural models may struggle with processing complex situations, while neural models find it difficult to handle entities that they have never seen. Such phenomena were not supported by the outcomes of~\citet{same-etal-2022-non} (cf. Appendix~\ref{sec:appendix_a}).


\paragraph{Rewriting.}

We quantify the results of the rewriting task by computing the rewriting rates (RR), defined as the proportion of rewritings over all REs generated by each model. The results of RR are also depicted in Table~\ref{tab:rating_webnlg} (raw counts of re-writings can be found in Appendix~\ref{sec:appendix_b}), from which we observe a similar trend to the previous results. Models that receive higher scores generally have lower RR. Neural models have low RR on seen data and high RR on unseen data while non-neural models have similar RR scores on seen and unseen data. 

\subsection{Relations between Ratings and Responses to the Tasks}

To test $\mathcal{H}_3$, we first tested how the referential success is correlated with clarity scores. 
To this end, we examine whether there is a significant positive correlation between clarity and the number of successful REs in a text (i.e., the sum of `YES' responses from our experiment and the number of \textsc{proper name}s) and the number of `YES' responses alone.\footnote{Recall that all REs in a text may be successful, but since they are all \textsc{proper name}s, participants were never asked about their success in our experiment.} We did linear regression tests (i.e., using each measure above to predict the clarity score) and reported the p-value as well as the slope ($\beta$) and the effect size computed using the coefficient of determination ($R^2$). The results show that both the number of `YES' responses ($\beta=.12, R^2=.0099, p<.001$) and the number of successful REs ($\beta=.057, R^2=.0048, p=.002$) positively correlate with clarity. This confirms $\mathcal{H}_3$.

Regarding the relation between the ratings and participants' rewritings of each RE ($\mathcal{H}_4$), we used a linear regression test to assess the correlation of the number of rewritings on each rating. We identified significant negative impacts on clarity ($\beta=-.34, R^2=.083, p<.001$), grammaticality ($\beta=-.36, R^2=.081, p<.001$), and coherence ($\beta=-.37, R^2=.089, p<.001$). These findings suggest an affirmative answer to $\mathcal{H}_4$.

As for the last hypothesis, $\mathcal{H}_5$, we compared scores from items where participants were familiar with the major entity to those where they were not. Mann Whitney U tests found no significant difference in clarity, grammaticality, coherence scores, or the number of rewritings. Therefore, we accepted the null hypothesis, suggesting that familiarity did not play a role in relation to this corpus.
\section{Further Observations} \label{sec:observation}

Additionally, we made the following observations while annotating the data.

\subsection{Additional Comments}

During the experiment, we received a few comments, which were supposed to focus on issues in the test items other than the contents of REs. These comments can be categorised into 4 types: (1) Potential ambiguities in the given text; (2) Ethical issues, for instance, ``\emph{Though the discourse is generally coherent, the pronouns are better neutralised (e.g., using `they' instead of `s/he')}''; (3) Inaccurate/inappropriate phrases other than REs; (4) Overall quality: quite a few comments suggest that the text as a whole is of low quality, for example, ``\emph{the language is unprofessional}'' or ``\emph{the discourse structure of the paragraph is bad}''. 

\subsection{Types of Rewritings}

We can use rewritings as a proxy to analyse the incorrectness and inappropriateness of REs. Thus, the rewritings can be roughly divided into two categories, as follows.

\paragraph{Correcting Errors in REs.} Some rewritings are about correcting errors in REs, including the following error types: (1) typo or grammatical error (``\emph{a admiral}'' $\to$ ``\emph{an admiral}''); (2) degeneration (``\emph{the Koc Koc}'' $\to$ ``\emph{The Koc}''); (3) definiteness (``\emph{AWH College}'' $\to$ ``\emph{the AWH College}''); (4) possessive (``\emph{Alan Frew}'' $\to$ ``\emph{Alan Frew's}''); (5) unknown referent: participants sometimes pointed out that given the RE and its context, it was not possible to infer which referent it refers to. This happened when there was serious referential ambiguity or the non-pronominal form of the referent had never appeared in the previous discourse; (6) incorrect referent: some rewritings changed the RE to refer to a completely different referent. It happened when (a) the context of the RE suggests that the RE at this position should refer to a different referent, (b) the RE is a pronoun and there is a mismatch in its surface form, and (c) the RE contradicts the common knowledge of the participant (e.g., we observed that multiple participants rewrote ``\emph{Elizabeth II}'' to ``\emph{Charles III}'').

\paragraph{Optimising REs.} 
Other rewritings aim at optimising the content of REs to make the whole discourse clearer and more coherent, including the following kinds: (1) referential form (``\emph{Alan Frew}'' $\to$ ``\emph{he}''); (2) punctuation (``\emph{the icebreaker Aleksey Chirikov}'' $\to$ ``\emph{the icebreaker, Aleksey Chirikov},''); (3) paraphrasing: some REs are paraphrased to be more readable (``\emph{the defender (football)}'' $\to$ ``\emph{the football defender}''); (4) elaboration/simplification: some REs were considered to be over-specificified or under-specified~\citep{chen2023varieties} and, thus, were simplified or elaborated in the rewritings; (5) non-RE: since our experiment did not limit participants to filling in only REs, rewritings could also be something other than referring expressions; (6) style (e.g., politeness: ``\emph{Alan Shepard}'' $\to$ ``\emph{Mr. Alan Shepard}''); 
(7) ethical issues (``\emph{he}'' $\to$ ``\emph{they}''). 
In a follow-up study, we plan to conduct a qualitative analysis of the rewritings and a detailed annotation of the rewriting types observed.

\subsection{The Context of an RE} 

As discussed, REG is highly context-dependent. Nonetheless, almost all computational REG models so far consider merely textual 
context.\footnote{Exceptions include, for example, \citet{cao-cheung-2019-referring}, who considered the knowledge about the major entity.} In this study, we observed multiple other kinds of contexts that also play important roles in human's use of RE. First, the optimality of REs in a discourse depends on each other. For example, participants sometimes rewrote REs to avoid duplication. 
However, the current setting of End2End REG models does not allow dependent production of REs. Second, the style of text contributes greatly to how elaborate the REs should be. In our case, since our data was in Wikipedia style, many participants thought the REs should be as elaborate as they could. As a consequence, for instance, they viewed ``\emph{Essex County}'' as an unsuccessful RE (although it is a distinguishing proper name) and rewrote it to ``\emph{Essex County, New York}''. Last, we observed quite a lot of cases where the background knowledge is influential. For example, participants from the United Kingdom often rewrite ``\emph{Elizabeth II}'' to ``\emph{Charles III}'' while those from the United States often rewrite ``\emph{Donald Trump}'' to ``\emph{Joe Biden}'' when referring to the leader of the country.


\section{Conclusion}

Focusing on some surprising results from \citet{same-etal-2022-non}, namely that REG models are indistinguishable from each other and from the REs in \textsc{webnlg} in a rating-based human evaluation, this paper has introduced a new type of intrinsic task-based REG evaluation, in which in parallel with rating, we designed two tasks: one asks participant's view about the referential success of each RE and the other asks participants to suggest a rewriting for each RE if possible. In this way, we had a better understanding of REG models' performance from different perspectives. Meanwhile, we confirmed our hope that accomplishing these meta-level tasks helped participants rate in a more reliable and discriminable way. 


This comprehensive evaluation suggests that, among the models we had tested, the machine learning based REG model performed the best on \textsc{webnlg}. Compared to the rule-based and neural models, it had a remarkably high rate of producing referentially successful REs. Additionally, it received the best clarity, grammaticality and coherence scores, and its generated content was rewritten the least frequently. 

We hope our design of the intrinsic task-based human evaluation protocol can serve as a reference for the evaluation of other NLG tasks that are complex enough so that simple evaluation cannot offer all the answers.
We also hope that the data from the experiment can help linguists understand the use of reference better and help computer scientists build better REG models.

In future, on the one hand, we plan to conduct more in-depth analyses of the issues we discussed in Section~\ref{sec:observation}, including qualitative and quantitative analyses of the rewritings and the factors that could affect 
the use of RE. On the other hand, our results revealed that neural models are good at processing seen data but not unseen data. This problem might be addressed by the most recent Large Language Models (LLMs).\footnote{We did not test LLMs as (1) our focus in this paper is not on seeking the best-performing REG models, and (2) this approach allows us to reuse materials from~\citet{same-etal-2022-non}, ensuring a fair comparison.} For this reason, in future research, we plan to apply mainstream LLMs to REG and assess these models using the protocol proposed and investigated
in this paper.

\section*{Ethical Considerations}

Two ethical considerations need to be noted here: One is that we recruited participants on Prolific. We only used their demographic information that is publicly available on Prolific. Our experiment does not collect any personal information. The other is that as a few participants reported, referring expressions in \textsc{webnlg} or produced by REG models might be gender-biased.

\section*{Limitations}


In this paper, we used the term ``neural model'' to refer to the NeuralREG models that we tested in this work and used the ``state-of-the-art'' to refer to the state-of-the-art when \citet{same-etal-2023-models} was carried out. As explained, our specific conclusions about neural models 
may not generalize to the most recent pre-trained Large Langauge Models. Second, this work considered only a simple dataset, namely, \textsc{webnlg}. It has been argued that \textsc{webnlg} is flawed as a tool for REG evaluation~\citep{chen2023neural} and the choice of the corpus would highly influence the evaluation results~\citep{same-etal-2023-models}. It is worth noting that, in this paper, we've only challenged the evaluation protocol used in previous studies; our findings 
do not focus on
the choice of evaluation corpus. Finally, in our experiment, we asked participants to only rewrite REs (rather than rewriting the entire text). This might somewhat decrease the ecological validity of the experiment as, normally, humans do not produce REs given 
linguistic contexts that have already been realized.

\bibliography{anthology,custom}

\clearpage
\appendix
\section{Results in \citet{same-etal-2022-non}} \label{sec:appendix_a}

Table~\ref{tab:human_webnlg_2022} shows the results for the models we examined in this work from~\citet{same-etal-2022-non}. Since, in this study, we tested fewer models, the factors of the Bonferroni correction would be different. Therefore, we re-did the significant testing and reported the results in Table~\ref{tab:human_webnlg_2022}. Additionally, we found a small error in the results in \citet{same-etal-2022-non}, which is also corrected in Table~\ref{tab:human_webnlg_2022}.

\section{Instruction of Our Experiment} \label{sec:instruction}

we explained the general goal to the participants and clarified what we expected them to do: 
\begin{quote}
    \emph{In this experiment, you will see 15 short paragraphs, each containing 2-3 sentences. We are particularly interested in the use of some of the expressions within these paragraphs. Accordingly, we will ask you to answer several questions about these expressions in the paragraphs. Given that language use can often be imperfect, the final question will ask you to suggest better alternatives for each expression in order to improve the clarity, grammaticality, and coherence of the paragraph.}
\end{quote}

\section{Raw Numbers from Our Experiment} \label{sec:appendix_b}

Table~\ref{tab:success_count} reports the raw number of answers to the questions about the referential success of REs. In addition to the findings in Section~\ref{sec:result}, the numbers here show that \texttt{ML} rarely produced REs that are not proper names when processing unseen data. Table~\ref{tab:rewriting_count} charts the number of rewritings. 

\begin{table}[h!]
\small
\centering
\begin{tabular}{lccc}
\toprule
Model & All & Seen & Unseen \\ \midrule
\texttt{RULE} & 603 & 341 & 262 \\
\texttt{ML} & 562 & 298 & 264 \\
\texttt{NREG-1} & 579 & 233 & 346 \\
\texttt{NREG-2} & 655 & 278 & 377 \\ \midrule
\texttt{Human} & 433 & 244 & 189 \\
\bottomrule
\end{tabular}
\caption{The number of rewritings.}
\label{tab:rewriting_count}
\end{table}
\begin{table*}[t!]
\small
\centering
\begin{tabular}{lccccccccc}
\toprule
 & \multicolumn{3}{c}{All} & \multicolumn{3}{c}{Seen} & \multicolumn{3}{c}{Unseen} \\\cmidrule(lr){2-4} \cmidrule(lr){5-7}\cmidrule(lr){8-10}
Model & Clarity & Grammar & Coherence & Clarity & Grammar & Coherence & Clarity & Grammar & Coherence \\ \midrule
\texttt{RULE} & 5.71$^{A}$ & 5.73$^{A}$ & 5.76$^{A}$ & 5.68$^{A}$ & 5.62$^{A}$ & 5.73$^{A}$ & 5.75$^{A}$ & 5.83$^{A}$~~~ & 5.79$^{A}$ \\
\texttt{ML} & 5.67$^{A}$ & 5.63$^{A}$ & 5.78$^{A}$ & 5.62$^{A}$ & 5.63$^{A}$& 5.73$^{A}$ & 5.72$^{A}$& 5.62$^{A,B}$ & 5.82$^{A}$ \\
\texttt{NREG-1} & 5.68$^{A}$ & 5.62$^{A}$ & 5.65$^{A}$ & 5.76$^{A}$ & 5.64$^{A}$ & 5.71$^{A}$ & 5.59$^{A}$ & 5.60$^{A,B}$ & 5.58$^{A}$ \\
\texttt{NREG-2} & 5.66$^{A}$ & 5.56$^{A}$ & 5.68$^{A}$ & 5.65$^{A}$ & 5.68$^{A}$ & 5.69$^{A}$ & 5.66$^{A}$ & 5.43$^{B}$~~~ & 5.67$^{A}$ \\\midrule
\texttt{Human} & 5.82$^{A}$ & 5.69$^{A}$ & 5.81$^{A}$ & 5.83$^{A}$ & 5.69$^{A}$ & 5.77$^{A}$ & 5.80$^{A}$ & 5.70$^{A,B}$ & 5.84$^{A}$ \\
\bottomrule
\end{tabular}
\caption{Human Evaluation Results from \citet{same-etal-2022-non} on the \textsc{webnlg} corpus. Rankings are determined by significance testing ($p < 0.01$; using Wilcoxon's signed-rank test with Bonferroni correction). 
Per column, results that have \emph{no} superscript letters in common are significantly different from each other.
}
\label{tab:human_webnlg_2022}
\end{table*}
\begin{table*}[t!]
\small
\centering
\begin{tabular}{lccccccccc}
\toprule
 & \multicolumn{3}{c}{All} & \multicolumn{3}{c}{Seen} & \multicolumn{3}{c}{Unseen} \\\cmidrule(lr){2-4} \cmidrule(lr){5-7}\cmidrule(lr){8-10}
Model & Yes & Maybe & No & Yes & Maybe & No & Yes & Maybe & No \\ \midrule
\texttt{RULE} & 131 & 9 & 4 & 65 & 4 & 3 & 66 & 5 & 1 \\
\texttt{ML} & 345 & 79 & 24 & 322 & 78 & 24 & 23 & 1 & 0 \\
\texttt{NREG-1} & 745 & 298 & 109 & 402 & 56 & 22 & 343 & 242 & 87 \\
\texttt{NREG-2} & 704 & 283 & 109 & 400 & 78 & 34 & 304 & 205 & 75 \\ \midrule
\texttt{Human} & 798 & 85 & 29 & 417 & 49 & 22 & 381 & 36 & 7 \\
\bottomrule
\end{tabular}
\caption{The count of each answer to the questions concerning the referential success of REs. }
\label{tab:success_count}
\end{table*}

\end{document}